\documentclass[11pt]{article}

\usepackage[final]{acl}

\usepackage{times}
\usepackage{latexsym}

\usepackage[T1]{fontenc}

\usepackage[utf8]{inputenc}

\usepackage{enumitem}
\setlist[itemize]{itemsep=2pt}

\usepackage{microtype}

\usepackage{inconsolata}

\usepackage{graphicx}

\usepackage{microtype}
\usepackage{booktabs}
\usepackage{array}
\usepackage{tabularx}
\usepackage{multirow}
\usepackage{pifont}
\newcommand{\cmark}{\ding{51}}%
\newcommand{\xmark}{\ding{55}}%
\usepackage{amsmath}
\usepackage{placeins}

\usepackage{ulem}

\usepackage{lineno}

\usepackage{caption}
\title{MedHal: a Synthetic Dataset for Medical Hallucination Detection}

\author{
\textbf{Fabrice Lamarche\textsuperscript{1,$\dagger$}\thanks{Correspondence: \texttt{fabrice.lamarche@mila.quebec}}},
\textbf{Gaya Mehenni\textsuperscript{1,$\dagger$}},
\textbf{Neshat Elhami Fard\textsuperscript{1}},
\textbf{Odette Rios-Ibacache\textsuperscript{2}},
\\
\textbf{Li Ming Wang\textsuperscript{3}},
\textbf{John Kildea\textsuperscript{2}},
\textbf{Amal Zouaq\textsuperscript{1}}
\\[1em]
\textsuperscript{1}LAMA-WeST Lab, Polytechnique Montr\'eal, Mila -- Quebec AI Institute \\
\textsuperscript{2}Medical Physics Unit -- RI-MUHC, McGill University \\
\textsuperscript{3}Department of Medicine, Division of Experimental Medicine, McGill University
\\[0.5em]
\textsuperscript{$\dagger$}Equal contribution
}

\begin{document}
\maketitle
\begin{abstract}
Hallucination, the generation of non factual content by AI systems, poses serious risks in medical contexts, where errors can directly affect patient outcomes. We present MedHal\footnote{MedHal is distributed under the Apache 2.0 license. Code, dataset and trained models can be found at \hyperlink{https://github.com/LAMA-WeST-Lab/MedHal}{https://github.com/LAMA-WeST-Lab/MedHal}}, a large-scale dataset specifically designed to assess capabilities and train models on the task of hallucination detection in medical texts. Current hallucination detection methods face significant limitations when applied to specialized domains like medicine, where they can have disastrous consequences. MedHal addresses this issue by incorporating diverse medical text sources and tasks covering both intrinsic and extrinsic hallucinations, and by providing a substantial volume of data samples suitable for training medical hallucination detection models. We demonstrate MedHal’s utility by training and evaluating a baseline medical hallucination detection model, showing improvements over general-purpose hallucination detection approaches. This resource enables more efficient evaluation and training of medical text generation systems while reducing reliance on costly expert review, potentially accelerating the development of medical AI research.
\end{abstract}

\section{Introduction}
Despite significant advancements in Natural Language Processing, Large Language Models (LLMs) are still prone to generating hallucinations \citep{Huang_2025_hallucination_survey}. This tendency poses a substantial obstacle to their application in multiple domains like medicine. Traditional methods based on Natural Language Inference \citep{kryscinski2020evaluating, goyal2021annotating} often fail to capture the nuances of factual correctness and are of limited accuracy. Although more recent approaches employing language models \citep{kim2024prometheus, min2023factscore} use contextual information for improved detection, these general approaches remain suboptimal in specialized domains \citep{deng2026beyond} where they fail to generalize.

In high-stakes contexts like medical text generation, the most reliable method for identifying hallucinated content remains expert human review \citep{searle_discharge_2023, hegselmann_medical_nodate}. However, this process is both time-consuming and resource-intensive, placing a considerable burden on medical professionals. Consequently, there is a clear need for a dedicated medical hallucination evaluation dataset, both to assess the performance of LLMs on hallucination detection tasks and to enable fine-tuning for improved hallucination detection capabilities. 

Currently, many medical datasets focus on single tasks such as Question Answering (QA) or Natural Language Inference (NLI), limiting their utility across the full range of medical text generation scenarios \citep{romanov_2018_lessons, chen_2024_detectingevaluatingmedicalhallucinations, hegselmann_medical_nodate, pandit2025medhallu}. Moreover, existing domain-specific hallucination datasets typically contain only a few hundred samples, rendering them insufficient for training or robustly evaluating LLMs \citep{hegselmann_medical_nodate, zuo2024medhallbench}. In this work, we propose MedHal, a dataset for the accurate detection of hallucinated content, which can ultimately facilitate broader adoption of medical AI systems.

\section{Related Work}
As defined by \citet{ji2023survey}, text generation hallucinations fall into two distinct categories: intrinsic hallucinations, which explicitly contradict the provided context, and extrinsic hallucinations, which introduce information that cannot be verified from the source material. While trustworthy medical hallucination detection must address both types, current methods in the medical domain face significant challenges due to the scarcity of comprehensive and diverse datasets.

Existing medical hallucination datasets often focus on specific tasks, are limited in size, or fail to cover all hallucination types. For instance, Med-Hallmark \citep{chen_2024_detectingevaluatingmedicalhallucinations}, Med-HVL \citep{yan_2024_medhvl}, and MedHallBench \citep{zuo2024medhallbench} are restricted to visual question-answering, evaluating models' ability to answer questions about medical images. Similarly, MedHalu \citep{agarwal_2024_medhaluhallucinationsresponseshealthcare}, Med-Halt \citep{pal_2023_medhaltmedicaldomainhallucination}, and MedHallu \citep{pandit2025medhallu} focus on text-based QA, assessing LLMs' ability to answer complex medical questions and detect internal inconsistencies. Consequently, these benchmarks cannot easily be adapted to general medical hallucination detection scenarios and lack the sufficient volume required to train dedicated detectors.

Furthermore, dataset scale remains a critical bottleneck. While the Hallucinations-MIMIC-DI dataset \citep{hegselmann_medical_nodate} provides detailed annotations of discharge summaries and BHC sections, its size is restricted to only 100 samples, rendering it insufficient for effective model training or evaluation. MHB \citep{lu2026mhb} takes a different approach by designing prompts specifically to elicit hallucinations from LLMs, a paradigm that is inherently unsuitable for training a hallucination detector. Finally, although large-scale datasets such as MedNLI \citep{romanov_2018_lessons} exist for medical Natural Language Inference (NLI) tasks, they rely on short, single-sentence premises and hypotheses. This structural limitation prevents them from capturing the complex medical reasoning and subtle hallucinations present in longer clinical contexts. For example, a typical MedNLI sample pairs a concise premise such as \textit{Labs were notable for Cr 1.7 (baseline 0.5 per old records) and lactate 2.4''} with a hypothesis such as \textit{Patient has normal Cr."}


Our work differs from traditional hallucination medical datasets in two key aspects: 
\begin{itemize}
    \item We utilize a greater variety of sources, such as clinical notes, clinical trials and medical questions, covering more complex contexts, across both \textit{intrinsic} and \textit{extrinsic} hallucinations.
    \item We create a large scale dataset and use it to train a medical evaluator that efficiently detects hallucinated content in out-of-distribution scenarios and human-annotated benchmarks.
\end{itemize}

\section{Methodology}
This study introduces MedHal, a novel dataset and benchmark designed for the evaluation of medical hallucination detection models. MedHal encompasses a diverse corpus of medical text, including clinical notes, research articles, and patient communication, annotated with various instances of factual inconsistencies and medical hallucinations. These hallucinations are generated through a suite of strategies tailored to individual task modalities, including answer substitution in question-answering (QA) and the introduction of conflicting statements in natural language inference (NLI). 
Subsequent sections detail the construction of MedHal, describe the associated benchmark, and present experimental results demonstrating its utility in evaluating medical hallucination detection models. The methodological approach involves the transformation of existing medical datasets across diverse tasks (QA, Summarization, NLI, Information Extraction) into a unified hallucination detection task. This is achieved by framing the task as the binary classification of a given statement as factual or non-factual, potentially conditioned on a provided context.

\begin{table*}[ht]
    \centering
    \small
    \resizebox{\textwidth}{!}{
    \begin{tabular}{|m{2cm}|m{7cm}|m{3cm}|}
        \toprule
        \textbf{Task}&  \textbf{Sample}& \textbf{Generated Statement}\\
        \midrule
        Information Extraction&   A 10-year-old girl first noted a swollen left knee and underwent repeated ... $\rightarrow$ \textbf{age: 10 years old}& The patient is 10 years old.\\
        \hline
        Summarization&   the large genotyping studies in the last decade have revolutionize genetic studies. our current ability to ... $\rightarrow$ \textbf{genetic admixture is a common caveat for ...}& genetic admixture is a common caveat for...\\
        \hline
        NLI&   Labs were notable for Cr 1.7 (baseline 0.5 per old records) and lactate 2.4. $\rightarrow$ \textbf{Patient has normal Cr}& Patient has normal Cr\\
        \hline
        QA&    Which of the following medications is most commonly used as treatment for type 2 diabetes?$\rightarrow$ \textbf{Metformin} & Metformin is most commonly used as treatment for type 2 diabetes.\\
        \bottomrule
    \end{tabular}
    }
    \caption{Example of samples used to generate statements for each task}
    \label{tab:examples_tasks}
\end{table*}

\subsection{Unified Task Formulation}
To construct the dataset, we establish a unified task formulation. Drawing inspiration from the NLI paradigm, each sample in the dataset is composed of a statement (analogous to the NLI hypothesis). This statement is classified as either factual or non-factual. The statement may be presented in the context of specific, intrinsic medical information (e.g., "\textit{This patient has suffered from a myocardial infarction}") or as a general, extrinsic assertion within the medical domain (e.g., "\textit{Aging causes an increase in blood pressure}"). For each statement, the dataset provides a binary label indicating its factuality and, in the case of non-factual statements, an explanation detailing the specific inconsistency. A statement is defined as factual if all information contained within it is verifiable through the provided context or by established medical knowledge. Each sample in MedHal is structured as follows:
\begin{itemize}
    \item Statement: A proposition to be classified as factual or non-factual.
    \item Context (optional): Contextual information relevant to the statement.
    \item Factual (Binary): A binary label indicating the factuality of the statement (Yes/No).
    \item Explanation: A textual explanation of the inconsistency for non-factual statements.
\end{itemize}
The following details the generation of samples for this unified task, based on datasets from question-answering, information extraction, NLI, and summarization tasks. Table \ref{tab:examples_tasks} shows examples of how correct  statements are generated from samples.

\vspace{3pt}
\noindent\textbf{Question Answering Dataset Transformation.}
\label{QA}
Question-answering (QA) datasets are structured around a query followed by a set of potential responses, including binary (yes/no/maybe) and multiple-choice (A, B, C, ...) formats. To generate factual samples from QA datasets, the question and its corresponding correct answer are transformed into a declarative statement using a large language model. Conversely, hallucinated samples are produced by pairing the question with incorrect answer options and subsequently converting these combinations into statements via the same LLM. Table \ref{tab:example_qa} in Appendix \ref{app:data_transformation_ex} shows an example of a multiple-choice question converted into a statement. For generation, a consistent prompt template is employed for both factual and hallucinated sample generation. The prompt is given in Appendix \ref{app:prompt_templates}. One-shot prompting is utilized to guide the LLM in accurately converting question-answer pairs into coherent statements. As for the explanation, certain QA datasets provide an explanation of why an answer is false. In these cases, we use the provided explanation for the non-factual statement. In other cases, we simply use the statement of the factual sample as the explanation. 

The QA task introduces extrinsic hallucinations because some original samples contain only answer choices with no accompanying context. After transformation into the MedHal format, these become standalone statements for which factuality must be assessed against current medical knowledge. For instance, the following statement : \textit{Increased level of IgM indicates recent vaccination.} is to be evaluated without context and only against current medical knowledge.

\vspace{3pt}
\noindent\textbf{Information Extraction Dataset Transformation.}
\label{IE}
Information extraction (IE) datasets pair source documents (e.g., clinical notes) with extracted text sequences representing specific document attributes. Factual samples use the source document as context and its extraction as the statement. Non-factual samples are generated by randomly swapping extractions of the same type across documents (e.g., medications exchanged between records). When extractions are key-value pairs, an LLM converts them into declarative statements (e.g., "visit motivation: Lower back pain" → "The patient's visit motivation is lower back pain") using the prompt in Appendix \ref{app:prompt_templates} with one-shot prompting. For non-factual samples, the explanation is the correct extraction from the same document and concept. Source documents are used without preprocessing. This task strictly contains intrinsic hallucinations since the contradicted information is always present in the context. An example of how a sample from an information extraction dataset is transformed is shown in Table \ref{tab:example_ie} in Appendix \ref{app:data_transformation_ex}

\vspace{3pt}
\noindent\textbf{Natural Language Inference Dataset Transformation.}
\label{NLI}
Given the inherent similarity between the proposed task and natural language inference (NLI), NLI datasets are directly adaptable to the MedHal dataset. We utilize NLI datasets specifically related to the medical domain. Factual samples are derived from hypotheses in entailment examples, while hallucinated samples are generated from hypotheses in contradiction examples. The premise component of the NLI example serves as the context for the corresponding statement. Samples resulting in a neutral label are ignored during dataset construction. The NLI datasets are not preprocessed beyond this filtering. Note that in this case, we do not have explanations for the non-factual samples.

This task also presents cases of extrinsic hallucinations, for samples where the inference necessitates external knowledge. For example, for a sample where the context is \textit{71yo female with known aortic stenosis.}, and the statement is \textit{The patient has normal cardiac function}, the factuality is to be established with regards to current medical knowledge.

\vspace{3pt}
\noindent\textbf{Summarization Dataset Transformation.}
\label{Sum}
For summarization datasets, factual samples are constructed using genuine summaries or sentences extracted from them. Hallucinated samples are generated by extracting a single sentence from the original summary and employing an LLM to modify it, introducing new information contradicting the source text. 
The modified hallucinated sentence is then reintegrated into the summary at its original position. The source text intended for summarization serves as the context, and the summary itself is treated as the statement. The explanation for a non-factual statement is simply the original sentence used to generate the hallucinated version. Sentences for hallucination generation are selected randomly, with a minimum length of 100 characters to ensure sufficient contextual information for the LLM. An example of how a sample is generated is shown in Table \ref{tab:example_sum} in Appendix \ref{app:prompt_templates}. The prompt used to generate self-contradictory sentences is shown in Appendix \ref{fig:prompt_template_sum}. The goal of this process is to evaluate if models can effectively detect fine-grained hallucinations in statements consisting of long sequences of text. This task also does not contain extrinsic hallucinations as the modified information is either contradicted or confirmed in the source text.

\begin{table*}[ht]
\centering
\resizebox{\textwidth}{!}{
\begin{tabular}{|l|l|c|l|l|l|l|}
\toprule
\textbf{Dataset} & \textbf{Task} & \textbf{S} & \textbf{Content Type} & \textbf{Source} & \textbf{\# Samples} & \textbf{\# Gen} \\ \midrule
MedMCQA & QA & \xmark & Medical Content & \citep{pal_2022_medmcqalargescalemultisubject} & 183,000 & 278,021\\ \hline
MedNLI & NLI & \xmark & Clinical Notes & \citep{romanov_2018_lessons} & 11,232 & 7,488\\ \hline
ACM & IE & \cmark & Clinical Notes & \citep{acm_2024} & 22,000 & 327,568\\ \hline
MedQA & QA & \xmark & Medical Content & \citep{jin_2020_diseasedoespatienthave} & 12,723 & 18,680\\ \hline
SumPubMed & Sum & \xmark & Clinical Trials & \citep{bharti_2020_pubmedsum} & 33,772 & 195,339\\
\bottomrule
\end{tabular}
}
\caption{Description of datasets, after filtering, used to create the MedHal benchmark (S is for Synthetic, \# Gen is the number of generated MedHal-format samples).}
\label{tab:dataset_description}
\end{table*}

\subsection{MedHal Dataset}
\subsubsection{Model}
For the creation of the MedHal dataset, we utilize the Llama-3.3-70B model \citep{grattafiori2024llama}, a language model of moderate size known for strong performance across language generation tasks. 
For all prompting strategies employed in the sample generation process, a one-shot learning approach was adopted to provide the model with examples of the desired output format prior to generation. See Section \ref{QA} for details on what is generated for each task.

\subsubsection{Dataset Creation}
To ensure a balanced benchmark, each factual sample is paired with a corresponding non-factual sample. Furthermore, the benchmark is constructed using a diverse set of datasets, encompassing all task modalities previously described, to ensure variability in document sources. Table \ref{tab:dataset_description} presents a comprehensive overview of the datasets utilized in the generation of this benchmark.

\subsubsection{Data Filtering}
\label{subsec:filtering}
To ensure data quality, we incorporated a filtering step where the Llama-3.3-70B model assessed the factuality of each generated statement, given its context. For tasks with subtle nuances, such as QA and summarization, the model was provided with the correct answer or the original sentence, respectively, to guide its judgment. For NLI tasks, it simply evaluated the statement's factuality. Samples were retained if the model's assessment matched the dataset's assigned label; otherwise, they were removed. This step serves as a validation layer in the data generation process, ensuring that any statements containing errors are identified and removed.

Manual observation of samples from the QA task revealed a tendency for the generation model to reformulate the task in its answer, leading to incorrect samples. 
To avoid this, we employed a second filtering step where statements containing certain trigger words like \texttt{`question'}, \texttt{`answer'} or \texttt{`statement'} are removed. See Appendix \ref{app:qa_filtering_keywords} for examples of faulty samples and the full keyword list. This was only done for the QA task since the others did not display this pattern.

Table \ref{tab:filtering} presents the percentage of samples retained for each dataset after this filtering process. A majority of samples were retained across most datasets, with the exception of ACM, which is the only synthetic dataset in our collection. MedNLI shows no removal since data was not modified for this task.

\subsubsection{MedHal Configurations}

The MedHal benchmark is distributed in several configurations to accommodate diverse research needs. The full dataset comprises a total of 828,974 samples. Additionally, we provide a length-filtered (LF) version, retaining only samples where the combined length of the context and statement is less than 30,000 characters. This specific length constraint is chosen to align with the 8,192 token context length common to 1-10B parameter models, reflecting our study's focus on the performance of smaller language models in medical settings.

Furthermore, we release a balanced version (BAL) of the dataset in terms of the tasks. In this configuration, the Question Answering (QA), Summarization, and Information Extraction (IE) tasks each contain approximately 150,000 samples. The NLI task, however, could not be balanced to this extent due to the limited number of available samples for that specific task. Table \ref{tab:dataset_distribution} details the number of samples and their respective proportions for each task across the different dataset versions. 

Importantly, to ensure a fair evaluation of our proposed benchmark, the dataset creation was performed separately on the different splits, with the initial training and testing sets from the original datasets retained. This allows for the independent evaluation of our trained models on the original test sets of the constituent datasets. The training portion of the MedHal dataset (LF + BAL) was then partitioned into training and validation sets following a 80-20 ratio, using stratified sampling to maintain consistent task proportions and class distributions across splits.

\begin{table*}[ht]
  \centering
  \begin{minipage}[t]{0.42\textwidth}
    \centering
    \begin{tabular}{lc}
      \hline
      \textbf{Dataset} & \textbf{\% Samples Kept} \\
      \hline
      SumPubMed  & 99.30 \\
      ACM        & 82.00 \\
      MedMCQA    & 90.76 \\
      MedQA      & 93.17 \\
      MedNLI     & 100.00 \\
      \hline
    \end{tabular}
    \caption{\textcolor{black}{Percentage of samples kept after both filtering steps.}}
    \label{tab:filtering}
    
  \end{minipage}
  \hfill
  \begin{minipage}[t]{0.54\textwidth}
    \centering
    \begin{tabular}{lccc}
      \hline
      \textbf{Task} & \textbf{Full} & \textbf{LF} & \textbf{LF+BAL} \\
      \hline
      Summarization         & 0.24 & 0.19 & 0.33 \\
      IE & 0.40 & 0.43 & 0.33 \\
      QA    & 0.36 & 0.36 & 0.33 \\
      NLI                   & 0.01 & 0.01 & 0.02 \\
      \hline
      Total samples         & 828,974 & 762,050 & 452,241 \\
      \hline
    \end{tabular}
    \caption{\textcolor{black}{Sample proportions per task (LF=Length Filtered, BAL=Balanced).}}
    \label{tab:dataset_distribution}
    
  \end{minipage}
\end{table*}

\subsection{Expert Validation}

The synthetic nature of the MedHal dataset necessitates a careful evaluation of its quality. To validate it, two of the co-authors, who are Ph.D. students in the medical field, conducted a manual annotation of a subset of the dataset. We randomly selected 100 examples from each task (50 for each factual and non-factual labels) to represent a statistically significant estimation of the data quality \footnote{More detail in Appendix \ref{app:sample_size}}. The annotators were tasked with validating the label, considering both the context and statement. A sample is considered valid if its label is judged as correct by the annotator. Annotation guidelines are presented in Appendix \ref{app:ann_guidelines}. Statistics describing the validity of each task is presented in Table \ref{tab:validation-results}. We observe very high validity for all tasks by both annotators. We also observe that one annotator identified many more invalid samples. We measured a raw agreement rate of 86\% and a \citep{gwet2001handbook}'s AC1 score of 0.83. This metric is recommended under high class imbalance \citep{zec2017high, wongpakaran2013comparison} which we both expect and measure here. These values indicate high agreement between raters on the quality of the dataset. Cases where either annotator marked a sample as invalid largely reflect situations where the label lacked sufficient specificity. For instance, statements being true at certain points in the clinical trajectory, but false at timepoints implied by the statement, or circumstances where statements were true in a narrow context but false if presented as a general claim. Given these results, we believe the MedHal dataset is of sufficient quality to allow for meaningful evaluation of models on the medical hallucination detection task.


\begin{table}[t]
\centering
\renewcommand{\arraystretch}{1.12}
\begin{tabularx}{\columnwidth}{@{}Xccc@{}}
\hline
\textbf{Task} & \textbf{A1} & \textbf{A2} & \textbf{Avg} \\
\hline
Common Knowledge       & 78.0 & 90.0 & 84.0 \\
Information Extraction & 86.0 & 98.0 & 92.0 \\
NLI                    & 81.0 & 97.0 & 89.0 \\
Summarization          & 100.0 & 100.0 & 100.0 \\
\hline
\textbf{Overall}       & \textbf{86.3} & \textbf{96.3} & \textbf{91.3} \\
\hline
\end{tabularx}
\caption{Expert validation results per task. Values are percentages. NLI = Natural Language Inference.}
\label{tab:validation-results}
\end{table}

\subsection{Data Leakage}
Concerns could be raised about potential prior exposure of models to the original datasets used in MedHal creation. First, we note that this would not impact the conclusions that we draw on the effects of MedHal finetuning. Regardless, we conduct two experiments assessing wether the evaluated models may have been exposed to portions of the MedHal test set during pretraining. These experiments, detailed in Appendix \ref{app:leakage}, show no evidence of systematic prior exposure.

\begin{table*}[ht!]
\centering
\color{black}
\begin{tabular}{|c|c|cc@{\hspace{0.4em}}cc|cc|}
\toprule
\multirow{2}{*}{\textbf{Type}}      & \multirow{2}{*}{\textbf{Model}} & \multicolumn{4}{c|}{\textbf{Hallucination Detection}}                              & \multicolumn{2}{c|}{\textbf{Explanation}}                             \\ \cline{3-6} \cline{7-8}
                           &                        & \textbf{H-F1}    & \textbf{NH-F1}    & \textbf{Mac-F1} & \textbf{$\Delta$F1} & \textbf{ECS} & \textbf{N} \\ \midrule
\multirow{6}{*}{General}   & Qwen3-8B               & 0.75 & 0.76 & 0.76 &            & 0.85 & 16,013 \\ \cline{2-8}
                           & \textsc{MedHal}-Qwen3-8B               & 0.89 & 0.88 & 0.89            & (+0.13) & 0.94 & 20,719\\ \cline{2-8}
                           & Llama-3-8B             & 0.66 & 0.74 & 0.70 &            &  0.83 & 18,724 \\ \cline{2-8}
                           & \textsc{MedHal}-Llama-3-8B & 0.90 & 0.90 & 0.90 & (+0.20)   & 0.94 & 21,055 \\ \cline{2-8}
                           & LLaMA3-70B             & 0.86 & 0.88 & 0.87 &            & 0.83 & 18,046 \\ \cline{2-8}
                           & GPT-OSS-20B            & 0.70 & 0.74 & 0.72 &            & 0.73 & 16,066 \\ \hline
\multirow{3}{*}{Medical}   & OpenBioLLM-8B          & 0.29 & 0.52 & 0.41 &            & 0.41  & 5,820 \\ \cline{2-8}
                           & \textsc{MedHal}-OpenBioLLM-8B & 0.60 & 0.74 & 0.67 & (+0.26) & 0.93 & 11,455 \\ \cline{2-8}
                           & OpenBioLLM-70B         & 0.64 & 0.68 & 0.66 &            & 0.55 & 9,309 \\ \cline{2-8}
                           & MedLlama3-8B & 0.61 & 0.74 & 0.67 &  & 0.61 & 11420 \\ \cline{2-8}
                           & \textsc{MedHal}-MedLlama3-8B & 0.89 & 0.89 & 0.89 & (+0.22) & 0.94 & 20,461 \\
                           \hline
\multirow{4}{*}{Evaluator} & Prometheus-2-8x7B      & 0.56 & 0.58 & 0.57 &            & 0.74 & 5,024 \\ \cline{2-8}
                           & HallOumi-8B            & 0.07 & 0.67 & 0.37 &            & 0.62 & 2,271 \\ \cline{2-8}
                           & \textsc{MedHal}-HallOumi-8B & 0.90 & 0.90 & 0.90 & \textbf{(+0.53)} & 0.93 & 20,735 \\ \hline
\multirow{4}{*}{Baseline}  & FactScore              & 0.60 & 0.38 & 0.49 &  & -- & -- \\ \cline{2-8}
                           & MedScore               & 0.63 & 0.45 & 0.54 &  & -- & -- \\ \cline{2-8}
                           & FaithLens              & 0.68 & 0.78 & 0.73 &  & -- & -- \\ \cline{2-8}
                           & PubMedBERT-NLI         & 0.61 & 0.54 & 0.58 &  & -- & -- \\ \bottomrule
\end{tabular}
\caption{Performance of models on MedHal's test set.
N refers to the number of samples where an explanation was extracted when prompting the model,
H is for `Hallucinated', NH for `Non-Hallucinated' and ECS for `Embedding Cosine Similarity', $\Delta$F1 is the improvement in Macro-F1 between the base and MedHal finetuned models.}
\label{tab:results_MedHal_general}
\end{table*}

\section{Experiments}

\subsection{General Evaluation}
To determine how well current LLMs detect medical hallucinations, we evaluated their performance on the MedHal test set. This assessment helps us identify which models are most adept at recognizing hallucinated content within a medical context and, more importantly, provides insights into whether specific fine-tuning strategies improve medical hallucination detection. 

We compare  the performance of general-purpose models (Qwen3-8B \citep{yang2025qwen3}, gpt-oss-20b \citep{agarwal2025gpt}, Llama-3-8B and 70B \citep{grattafiori2024llama}), medical models fine-tuned on general medical datasets (OpenBioLLM-8B and 70B \citep{OpenBioLLMs}, MedLlama-3-8B \citep{medllama}), and evaluator models fine-tuned to judge answers of other LLMs (Prometheus-2-8x7B \citep{kim2024prometheus} and  HallOumi-8B \citep{oumi2025HallOumi}). The specific model checkpoints are listed in Appendix \ref{app:checkpoints}. We specifically chose the OpenBioLLM-8B, MedLlama-3-8B and HallOumi-8B models because their base models is the Llama-3-8B model we also used, which allows comparison of initial fine-tuning strategies. For general-purpose and medical models, we used the prompt template detailed in Figure \ref{fig:prompt_template_eval} with 2-shot prompting. For Prometheus-2-8x7B and HallOumi-8B, we adhered to the prompt formats recommended by their original authors, as these are optimized for their fine-tuned performance. Model-specific sampling arguments are detailed in Appendix \ref{app:checkpoints}.

We also include hallucination detection baselines in our evaluation : 
\begin{itemize}
    \item \textbf{FactScore} \citep{min2023factscore} : This method typically decomposes the statement to verify and checks each atom with knowledge from a knowledge base. Here we pose the context as the knowledge base, and use the open-source Qwen3-8B model for decomposition, using the default prompt from FactScore.
    \item \textbf{MedScore} \citep{huang-etal-2026-medscore} : MedScore is a medical-focused evolution of the FactScore method, where we again use the input context as the knowledge base for atom verification and use the same model for decomposition.
    \item \textbf{FaithLens} \citep{si-etal-2026-faithlens} : This general-purpose detection method shows state-of-the-art results on multiple benchmarks, using a fine-tuned variant of the Llama-3.1-8B-Instruct Model \citep{grattafiori2024llama}.
    \item \textbf{PubMedBERT-NLI} \citep{deka-etal-2023-multiple} : PubMedBERT-NLI is a NLI fine-tuned variant of the PubMedBERT model on MNLI and MedNLI.

\end{itemize}

The metric we use to evaluate models'  hallucination detection capabilities is the Macro F1-score, the average F1 score for both the hallucination (H) and non-hallucination (N-H) class. This metric is chosen because it penalizes poor performance on either the hallucinated or non-hallucinated samples and ensures a balanced evaluation. Results are presented in Table \ref{tab:results_MedHal_general}. This table also presents Explanation Metrics, which measure how closely the models'  generated explanations resemble those in our reference standard. To measure this, we compute the cosine similarity between embeddings of the reference and generated explanations, using the \texttt{Qwen3-Embedding-8B} \citep{qwen3embedding} model. While this does not allow an exact measurement of the explanation \textit{correctness}, it still provides an interesting signal into the explanation similarity to the dataset explanation. To ensure a valid comparison, this is only measured on samples where both the model's prediction and the ground truth label indicate a non-factual statement.

\subsection{MedHal fine-tuning}
We fine-tuned the above-referenced models on the MedHal training set to assess the impact of specialized training on their performance. Our main goal was to establish whether fine-tuning a medical model specifically for medical hallucination detection, or specializing a general hallucination detection model on medical data, would yield better results. We also investigated whether simply fine-tuning a general-purpose model could achieve similar performance, potentially bypassing the need for more specialized initial training. The prompt used during training is shown in Figure \ref{fig:prompt_template_train}. All models were fine-tuned using the same QLora configuration\footnote{For details on training setup, see Appendix \ref{app:training_config}.}. We report the same metrics detailed in Table \ref{tab:results_MedHal_general}, along with the difference in F1 score between the fine-tuned and non-fine-tuned versions of each model. MedHal finetuned models are tested under the zero-shot setting.

We also conduct an ablation experiment where we fine-tuned the Llama-3-8B model on each task individually, showing that they
contribute uniquely to improving model performance at hallucination detection. Results and more details are provided in Appendix \ref{task_ablation}.

\subsection{Downstream Performance}
We assess the utility of our MedHal dataset by evaluating the downstream task performance  of models fine-tuned on MedHal on three human-annotated hallucination detection datasets. Performance is reported in Table \ref{tab:downstream_perf}.

Hall-Gen-DI,  \citep{hegselmann_medical_nodate} is derived from MIMIC-III and contains summaries of patient notes generated by frontier models, which were then annotated for factual inconsistencies by medical students. We consider a summary as non-factual if it contains any of the eleven types of factual inconsistencies in the dataset, including 'Fact Contradicted', 'Unsupported Medication', or 'Unsupported Procedure'. Next, we use the Natural Hallucination (NH) dataset derived from ACI-Bench \citep{BN2024fact}. The NH dataset is used to evaluate hallucination detection in medical text summarization and is built on ACI-Bench clinician-patient conversation transcripts and their corresponding clinical summaries.
Hallucination is defined as information in the summary that is not present in or is misinterpreted from the source text.
Based on the LLM-generated summaries, clinical experts annotated each statement into four categories: Hallucination, Inference, Misunderstanding, and No Factual Error. For our unified format, we use the conversation ``context'', and a single statement from the LLM-generated summary as ``statement''. We construct the evaluation subset by selecting up to 88 no-error examples and 82 high-severity hallucination examples, yielding 170 samples in total. Finally, we use the Medec-UW \citep{abacha2025medec} test set, a benchmark for medical error detection and correction in clinical notes where errors were manually introduced into clinical notes by medical students. In our setting, the non-factual case corresponds to a sentence containing an introduced error in diagnosis, management, treatment, pharmacotherapy, or causal organism. Factual cases are random sentences from the clinical notes. This yields 328 evaluation samples in total from the Medec-UW test split, balanced across labels.

Although these dataset are all relatively small, their recent release makes it highly unlikely that the evaluated models were exposed to them during training. This mitigates the risk of data contamination ensuring a more objective evaluation. A summary of the downstream evaluation datasets and their transformation is shown in Appendix \ref{app:data_transformation_ex}.



\begin{table*}[ht]
\centering
\resizebox{\textwidth}{!}{%
\begin{tabular}{|c|l|ccc|}
\toprule
\textbf{Type} & \textbf{Model} & \textbf{ACI-NH} & \textbf{Medec-UW} & \textbf{Hall-Gen-DI} \\
\midrule
\multirow{6}{*}{General LLM}
 & Llama3-8B                & 0.54 $\pm$ 0.07 & 0.91 $\pm$ 0.03 & 0.50 $\pm$ 0.05 \\
 & \textbf{\textsc{MedHal}-Llama3-8B} & 0.52 $\pm$ 0.07 & 0.98 $\pm$ 0.01 & 0.63 $\pm$ 0.06 \\
 & Llama3-70B               & 0.63 $\pm$ 0.07 & 0.98 $\pm$ 0.01 & 0.51 $\pm$ 0.05 \\
 & Qwen3-8B                 & 0.62 $\pm$ 0.07 & 0.95 $\pm$ 0.02 & 0.65 $\pm$ 0.06 \\
 & \textbf{\textsc{MedHal}-Qwen3-8B}  & 0.63 $\pm$ 0.07 & 0.98 $\pm$ 0.01 & 0.63 $\pm$ 0.06 \\
 & GPT-OSS-20B              & 0.46 $\pm$ 0.08 & 0.64 $\pm$ 0.05 & 0.55 $\pm$ 0.06 \\
\midrule
\multirow{5}{*}{Medical LLM}
 & OpenBioLLM-8B                & 0.34 $\pm$ 0.07 & 0.79 $\pm$ 0.04 & 0.54 $\pm$ 0.06 \\
 & \textbf{\textsc{MedHal}-OpenBioLLM-8B} & 0.37 $\pm$ 0.07 & 0.90 $\pm$ 0.03 & 0.27 $\pm$ 0.03 \\
 & OpenBioLLM-70B               & 0.40 $\pm$ 0.07 & 0.95 $\pm$ 0.02 & 0.67 $\pm$ 0.06 \\
 & MedLlama3-8B                 & 0.53 $\pm$ 0.07 & \textbf{0.99 $\pm$ 0.01} & 0.62 $\pm$ 0.06 \\
 & \textbf{\textsc{MedHal}-MedLlama3-8B}  & 0.53 $\pm$ 0.07 & 0.89 $\pm$ 0.03 & 0.62 $\pm$ 0.06 \\
\midrule
\multirow{3}{*}{Evaluator LLM}
 & HallOumi-8B                & 0.44 $\pm$ 0.08 & 0.83 $\pm$ 0.04 & 0.46 $\pm$ 0.06 \\
 & \textbf{\textsc{MedHal}-HallOumi-8B} & 0.64 $\pm$ 0.07 & \uline{0.98 $\pm$ 0.01} & \textbf{0.69 $\pm$ 0.05} \\
 & Prometheus-2-8x7B          & 0.41 $\pm$ 0.08 & 0.78 $\pm$ 0.04 & 0.42 $\pm$ 0.05 \\
\midrule
\multirow{4}{*}{Baseline}
 & FactScore          & 0.62 & 0.71 & 0.45 \\
 & MedScore           & 0.45 & 0.77 & 0.48 \\
 & FaithLens & \textbf{0.70} & 0.91 & 0.41 \\
 & PubMedBERT-NLI     & 0.41 & 0.68 & 0.43 \\
\bottomrule
\end{tabular}%
}
\caption{Downstream performance (Macro-F1) on the three human-annotated hallucination detection datasets, with $\pm$ 95\% stratified bootstraped confidence interval over resampling of the datasets.}
\label{tab:downstream_perf}
\end{table*}

\section{Discussion}
\subsection{MedHal performance}

Our evaluation reveals several key insights into the performance of various models in detecting and explaining medical hallucinations. We first observe the large improvements that finetuning on MedHal provides, across all tested models, with the largest improvement observed for the evaluation-trained variant, Halloumi-8B, which achieves a 53-point increase in Macro-F1. We hypothesize this is because the model had evaluation capabilities that were unlocked by the medical specificity of the MedHal finetuning. The MedHal-Llama3-8B and MedHal-Halloumi-8B models both obtain the best scores on this evaluation followed closely by the MedHal-Qwen3-8B and MedHal-MedLlama3-8B variants. In contrast, the MedHal-OpenBioLLM-8B model does not reach the same performance, indicating that the extensive medical finetuning this model underwent might not allow it to adapt to the hallucination detection task. 

Another (expected) finding is the clear superiority of bigger and newer models compared to the base Llama3-8B model. Excluding the MedHal fine-tuned models, the larger  base Llama3-70B model and newer Qwen3-8B and gpt-oss-20b models obtain the top three LLM scores. They are, however, all surpassed by every MedHal fine-tuned variant, except the MedHal-OpenBioLLM-8B model, highlighting the benefits of our training process. We also observe that the hallucination detection baselines are inferior to some of the latest and largest LLMs, Qwen3-8B and Llama3-70B.  
When assessing the similarity of explanations to the reference explanations for non-factual content, we also observe large improvements after the fine-tuning process.

\subsection{Evaluation on Downstream Datasets}
The results on the three downstream, human-annotated datasets, shown in Table \ref{tab:downstream_perf}, highlight the benefits of MedHal fine-tuning. The MedHal-Halloumi-8B model achieves statistically significant improvements across all three datasets, increasing Macro-F1 scores by 15 to 23 points and achieving the best performance on two of the three datasets. The Llama3-8B model also gets significant performance improvements on two of the datasets, surpassing or achieving similar results to the larger Llama3-70B model. The FaithLens baseline is the strongest of the four and obtains the best performance on one dataset, but is surpassed both in average score and average ranking by these two MedHal fine-tuned variants. Across the three downstream datasets, the MedHal-Halloumi-8B model obtains the best performance overall. On the other hand, the medical OpenBioLLM-8B and MedLlama3-8B models and their MedHal finetuned variants somewhat struggle, with the finetuning process only improving performance on one of the datasets for both models. We hypothesize again that this is because the MedHal training interfered with the significant medical instruction tuning that the medical models already underwent. 
The MedHal variants are evaluated under the zero-shot setting, while the base models rely on two-shot in-context learning, making these results even more noteworthy: MedHal fine-tuned models generalize better to unseen datasets than their base counterparts do with explicit examples provided at inference time.
Overall, MedHal finetuning allows smaller and older models based on the Llama3-8B architecture to compete and perform better than larger and newer models (Llama3-70B, Qwen3-8B, gpt-oss-20b) on out-of-distribution, human-annotated medical hallucination detection datasets.

\section*{Conclusion}
In this study, we introduced MedHal, the first large-scale dataset designed to specifically address the challenge of hallucination detection across diverse clinical tasks, including summarization, natural language inference, information extraction, and question answering. We evaluated a range of medical and general-purpose models on this dataset, providing a comprehensive benchmark for assessing their performance in this critical domain. Furthermore, we fine-tuned multiple models on MedHal to evaluate the impact of specialized pre-training on medical hallucination detection performance. We show that models fine-tuned on MedHal surpass performance of larger and newer models on downstream, human-annotated hallucination detection tasks. Overall, MedHal represents a significant resource for addressing the limitations of existing hallucination detection methods in the medical field.

\section*{Limitations}
The main limitation of our work is  that we rely on generated content. Despite our expert validation step, clinical relevance and factuality of every sample cannot be fully guaranteed. An other improvement would be to incorporate more complex multi-hop reasoning hallucination detection tasks into the dataset. This would allow us to identify error patterns and to improve hallucination detection models.

The source datasets used in the construction of the MedHal corpus represent an additional limitation worth acknowledging. While all datasets are either anonymized to protect personal information or derived from publicly available material, they may contain more subtle forms of bias. Individual samples may make medical claims that have since been contradicted by more recent evidence, posing potential harms to specific patient populations. Considered collectively, these datasets may also perpetuate harm through systematic under-representation of certain demographic groups. A more thorough investigation into the distribution of content would therefore be necessary to fully ensure fairness and adequate representation.

Finally, this resource is only available in English and augmenting it with other languages would be beneficial. This is left for future work. 

\section*{Acknowledgments}
This project was undertaken thanks to funding from the FRQS (Fonds de Recherche du Québec – Santé) Chair in Artificial Intelligence and Digital Health. We also gratefully acknowledge the financial support provided by IVADO and the Canada First Research Excellence Fund. This research benefited from the computational resources and GPU infrastructure provided by the Digital Research Alliance of Canada (DRAC) and Calcul Québec.

\bibliography{custom_pruned}

\newpage

\appendix
\section*{Appendix}\label{sec:appendix}

\setlength{\textfloatsep}{8pt plus 2pt minus 2pt}
\setlength{\floatsep}{6pt plus 2pt minus 2pt}
\setlength{\intextsep}{8pt plus 2pt minus 2pt}

\section{Data Transformation Examples}\label{app:data_transformation_ex}

To illustrate how each source dataset was converted into the unified factuality format, we provide a worked example for each of the tasks. Table~\ref{tab:example_qa} shows an example from the Question Answering task, where a QA pair is rewritten as a factual or non-factual statement. Table~\ref{tab:example_ie} shows the equivalent transformation for Information Extraction, where a structured field extracted from a clinical note is converted into a natural-language sentence, with the non-factual variant substituting a mismatched value from another sample. Table~\ref{tab:example_sum} shows the Summarization case, where a generated summary sentence is either preserved as-is (factual) or perturbed with a deliberate inaccuracy (non-factual). Finally, Table~\ref{tab:downstream_datasets_unified} summarizes how each of the three downstream evaluation datasets (ACI-NH, Medec-UW, and Hall-Gen-DI) was mapped onto this same unified \texttt{context}/\texttt{statement}/\texttt{label}/\texttt{explanation} schema.

\section{Prompt templates}
\label{app:prompt_templates}
Figures \ref{fig:prompt_template_qa}, \ref{fig:prompt_template_ie}, \ref{fig:prompt_template_sum}, \ref{fig:prompt_template_eval} and \ref{fig:prompt_template_train} show the different prompt templates used in MedHal creation, training and evaluation.

\section{QA Filtering}
\label{app:qa_filtering_keywords}

Table \ref{tab:qa_filtering_ex} shows examples of faulty samples that were removed by this filtering step.

\begin{table}[h]
\centering
\begin{tabular}{|c|}
\hline
\textbf{Keywords} \\
\hline
question \\
answer \\
statement \\
however \\
above \\
shown \\
\hline
\end{tabular}
\caption{List of Keywords used for QA filtering}
\label{tab:qa_filtering}
\end{table}

\begin{table*}[t]
\centering
\begin{tabular}{|m{3cm}|m{3cm}|m{2cm}|m{3cm}|}
\toprule
\textbf{Sample Type} & \textbf{Question} & \textbf{Answer} & \textbf{Generated Statement} \\ \midrule
Factual & \multirow{2}{=}{Which of the following medications is most commonly used as first-line treatment for newly diagnosed type 2 diabetes mellitus in patients without contraindications?} & Metformin & Metformin is most commonly used as first-line treatment for newly diagnosed type 2 diabetes mellitus in patients without contraindications. \\ \cline{1-1} \cline{3-4}
Non-Factual & & Insulin & Insulin is most commonly used as first-line treatment for newly diagnosed type 2 diabetes mellitus in patients without contraindications. \\ \bottomrule
\end{tabular}
\caption{Example of Question-Answering Dataset Transformation}
\label{tab:example_qa}
\end{table*}

\begin{table*}[t]
\centering
\begin{tabularx}{\textwidth}{|l| >{\raggedright\arraybackslash}X | l | >{\raggedright\arraybackslash}X | >{\raggedright\arraybackslash}X |}
\toprule
Sample Type & Source Document & Extraction & Statement & Explanation \\
\midrule
Factual & A 10-year-old girl first noted a swollen left knee and underwent repeated arthrocentesis... & age: 10 years old & The patient is 10 years old & - \\
\hline
Non-Factual & A 10-year-old girl first noted a swollen left knee and underwent repeated arthrocentesis... & age: 16 years old & The patient is 16 years old & The patient is 10 years old \\
\bottomrule
\end{tabularx}
\caption{Example of Information Extraction Dataset Transformation (the extraction from the non-factual statement is taken from another original sample)}
\label{tab:example_ie}
\end{table*}

\begin{table*}[h]
\centering
\begin{tabularx}{\textwidth}{|l| >{\raggedright\arraybackslash}X | >{\raggedright\arraybackslash}X | >{\raggedright\arraybackslash}X | >{\raggedright\arraybackslash}X |}
\toprule
Sample Type & Source Document & Summary & Statement & Explanation \\
\midrule
Factual & a central feature in the maturation of hearing is a transition in the electrical signature of cochlear hair cells from spontaneous calcium... & cochlear hair cells are high-frequency sensory receptors... & cochlear hair cells are high-frequency sensory receptors... & - \\
\cline{1-1} \cline{4-5}
Non-Factual & & & cochlear hair cells are \textbf{low-frequency} sensory receptors... & According to the source document, cochlear hair cells are high-frequency sensory receptors... \\
\bottomrule
\end{tabularx}
\caption{Example of Summarization Dataset Transformation}
\label{tab:example_sum}
\end{table*}

\begin{table*}[t]
\centering
\footnotesize
\setlength{\tabcolsep}{4pt}
\renewcommand{\arraystretch}{1.05}
\begin{tabularx}{\textwidth}{|p{1.35cm}|p{2.3cm}|p{2.55cm}|>{\raggedright\arraybackslash}X|>{\raggedright\arraybackslash}X|p{1.45cm}|}
\toprule
\textbf{Dataset} & \textbf{Input} & \textbf{Output / task} & \textbf{Non-factuality / hallucination} & \textbf{Unified transformation} & \textbf{Final samples} \\
\midrule

\textbf{ACI-NH}
& Clinician--patient transcript and LLM-generated summary statements
& Factuality classification of each summary statement
& Summary information that is not present in, or is misinterpreted from, the source transcript. \emph{Hallucination}, \emph{Inference}, and \emph{Misunderstanding} are mapped to non-factual; \emph{No Factual Error} is factual.
& \texttt{context}: transcript;
\texttt{statement}: each summary statement;
\texttt{label}: YES for \emph{No Factual Error}, NO otherwise;
\texttt{explanation}: annotator notes when available.
& 170 \newline (88 YES, 82 NO) \\

\midrule

\textbf{Medec-UW}
& Clinical note text
& Error flag prediction, error sentence extraction, and correction generation
& An inaccurate medical statement introduced into the note, e.g., an error in diagnosis, management, treatment, pharmacotherapy, or causal organism; in the Medec-UW subset, injected errors are designed to contradict at least two other parts of the note.
& In our unified transformation, we remove one target sentence from the note;
\texttt{context}: remaining note text;
\texttt{statement}: removed sentence; we use the annotated error sentence when available, otherwise a randomly selected sentence;
\texttt{label}: YES for \texttt{Error Flag}=0, NO for \texttt{Error Flag}=1;
\texttt{explanation}: corrected sentence when available, otherwise ``The statement is factual.''
& 328 \newline (164 YES, 164 NO) \\

\midrule

\textbf{Hall-Gen DI}
& Original abstract and LLM-generated summary
& Binary factual consistency classification
& Summary content not contained in, or not logically derivable from, the abstract; a summary is consistent only if all its information is supported by the source.
& \texttt{context}: original abstract;
\texttt{statement}: generated summary;
\texttt{label}: YES for score 3, NO otherwise;
\texttt{explanation}: expert inconsistency rationale when available, otherwise ``The statement is factual.''
& 248 \newline (132 YES, 116 NO) \\

\bottomrule
\end{tabularx}
\caption{Downstream evaluation datasets and their transformation into the unified factuality format.}
\label{tab:downstream_datasets_unified}
\end{table*}

\begin{figure*}[h]
\centering
\fbox{%
  \begin{minipage}{0.8\textwidth}
    \smallskip
    \noindent
    \textbf{system:} Given a question and an answer, your role is to transform the question into a statement by incorporating the answer with it. Do not add any details that is not mentioned in the question or the answer.
    \smallskip
    \noindent
    \textbf{user:} Question: Which of the following agents is most commonly associated with recurrent meningitis due to CSF leaks?

    Answer: Pneumococci
    \smallskip
    \noindent
    \textbf{assistant:} Pneumococci is most commonly associated with recurrent meningitis due to CSF leaks
    \smallskip
    \noindent
    \textbf{user:} Question: [question]

    Answer: [answer]
  \end{minipage}
}
\caption{Prompt template used for the QA task}
\label{fig:prompt_template_qa}
\end{figure*}

\begin{figure*}[h]
\centering
\fbox{%
  \begin{minipage}{0.8\textwidth}
    \smallskip
    \noindent
    \textbf{system:} You are tasked with transforming structured medical data into natural language statements about a patient. Each input will contain 4 elements:

    - concept: The type of information being described (e.g., dosage, age, symptoms)

    - value: The specific information or measurement

    - category: The broad medical category this information belongs to (e.g., treatment, patient information, symptoms)

    - concept\_reference: The specific element that the value refers to (e.g., a specific medication, a specific symptom)

    Your task is to generate a clear, grammatically correct sentence that conveys this information in a medical context. Follow these rules:

    1. Use appropriate verbs based on the concept:
    - For treatments: 'takes', 'receives', 'is prescribed'
    - For symptoms: 'experiences', 'reports', 'presents with'
    - For measurements/states: 'is', 'has', 'shows'
    - For time-related concepts: 'has been', 'started', 'continues'

    2. Incorporate the concept\_reference when it adds clarity

    3. Use present tense

    4. Maintain medical terminology as provided

    5. When the concept\_reference is 'None' or does not add clarity, don't include it in the statement

    6. The statement should be a single sentence.

    Do not include any other information in the statement aside from the concept and the extraction. Only output the statement and nothing else.
    \smallskip
    \noindent
    \textbf{user:}
    category: patient medical history

    value: History of left elbow arthrodesis performed for posttraumatic arthritis at the age of 18,

    concept\_reference: None
    \smallskip
    \noindent
    \textbf{assistant:} The patient underwent left elbow arthrodesis as a treatment for posttraumatic arthritis when they were 18 years old.
    \smallskip
    \noindent
    \textbf{user:} category: [category]

    value: [value]

    concept\_reference: [concept\_reference]
  \end{minipage}
}
\caption{Prompt template used for the IE task (for more information, see \citep{acm_2024})}
\label{fig:prompt_template_ie}
\end{figure*}

\begin{figure*}[h]
\centering
\fbox{%
  \begin{minipage}{0.8\textwidth}
    \smallskip
    \noindent
    \textbf{user:} You will be given a text and a sentence that was extracted from the text.
    Your task is to transform the sentence by introducing a deliberate inaccuracy. Strategies can include:

    - Changing numerical values

    - Inverting the meaning

    - Using antonyms

    - Negating the original statement

    Text: [text]
    Sentence: [sentence]

    Ensure the new sentence remains grammatically correct but semantically different from the original. Only output the transformed sentence without any additional text.
  \end{minipage}
}
\caption{Prompt template used for the summarization task}
\label{fig:prompt_template_sum}
\end{figure*}

\begin{figure*}[h]
\centering
\fbox{%
  \begin{minipage}{0.8\textwidth}
    \smallskip
    \noindent
    \#\#\# Task Description

    - You will evaluate whether a medical statement is factually accurate.

    - The statement may reference a provided context.

    - Respond with "YES" if the statement is factually correct or "NO" if it contains inaccuracies.

    - In order to answer YES, everything in the statement must be supported by the context.

    - In order to answer NO, there must be at least one piece of information in the statement that is not supported by the context.

    - You must also provide an explanation of why you think the statement is factual or not. If it is factual, put "The statement is factual" as your explanation.

    - Your answer should follow the following format :

    Factual: [YES/NO]

    Explanation: [Your explanation]
    \smallskip
    \#\#\# Context

    [context]
    \smallskip
    \#\#\# Statement

    [statement]
  \end{minipage}
}
\caption{Prompt template used for evaluating models}
\label{fig:prompt_template_eval}
\end{figure*}

\begin{figure*}[h]
\centering
\fbox{%
  \begin{minipage}{0.8\textwidth}
    \smallskip
    \noindent
    \#\#\# Task Description

    - You will evaluate whether a medical statement is factually accurate.

    - The statement may reference a provided context.

    - Respond with "YES" if the statement is factually correct or "NO" if it contains inaccuracies.

    - In order to answer YES, everything in the statement must be supported by the context.

    - In order to answer NO, there must be at least one piece of information in the statement that is not supported by the context.

    - You must also provide an explanation of why you think the statement is factual or not. If it is factual, put "The statement is factual" as your explanation.
    \smallskip
    \#\#\# Context

    [context]
    \smallskip
    \#\#\# Statement

    [statement]
  \end{minipage}
}
\caption{Prompt template used for training models}
\label{fig:prompt_template_train}
\end{figure*}

\begin{table*}[h]
\centering
\begin{tabular}{|p{14cm}|}
\hline
\textbf{Statement} \\
\hline
Serine, not threonine, is typically involved in the synthesis of ceramide; however, according to the given answer: Threonine is involved in the synthesis of ceramide. \\
\hline
Vitamin K should be administered to reverse the effects of the heparin, however protamine is typically used to reverse heparin, but since the answer is Vitamin K, the statement will be:
Vitamin K should be administered to reverse the effects of the heparin.
 \\
\hline
The normal value of ankle brachial index is 0.8 or greater, however since 0.8 is given as the answer, the statement would be:
The normal value of ankle brachial index is 0.8, but to be more precise:
The normal value of ankle brachial index is at least 0.8, but since the exact wording of the answer is 0.8, the statement is:
The normal value of ankle brachial index is 0.8.
\\
\hline
\end{tabular}
\caption{Example of faulty Statements from the QA task.}
\label{tab:qa_filtering_ex}
\end{table*}

Table \ref{tab:qa_filtering} lists all keywords that were used for filtering the QA samples in Section \ref{subsec:filtering}.

\section{Model Checkpoints and Sampling Arguments}
\label{app:checkpoints}

Table \ref{tab:checkpoints} lists all used model checkpoints and their download links. Table \ref{tab:samplng} lists the model-specific sampling arguments that were used for evaluation. These were chosen individually according to the recommended settings. These settings were all found on the Hugging Face model page of the respective model, excepted for the GPT-OSS-20B model, where the recommended sampling arguments are stated by OpenAI representative in a discussion (\url{https://huggingface.co/openai/gpt-oss-120b/discussions/21}), and the Prometheus model for which the sampling arguments are stated in the introductory paper \citep{kim2024prometheus}. No recommended settings were found for the MedLlama3-8B and HallOumi-8B models, they were thus ran with the following default values:

\begin{itemize}
    \item Temperature = 0.0
    \item Top P = 1.0
    \item Top K = -1
    \item Min P = 0.0
\end{itemize}

These were also the sampling arguments used for all MedHal fine-tuned variants of the models.

Inference was conducted using the vLLM framework \citep{kwon2023efficient} with the computational setup described in Section \ref{app:compute_setup}.

\begin{table*}[h]
\centering
\resizebox{\textwidth}{!}{
\begin{tabular}{ll}
\hline
\textbf{Checkpoint} & \textbf{Link} \\
\hline
Qwen3-8B \citep{yang2025qwen3} & \url{https://huggingface.co/Qwen/Qwen3-8B} \\
Llama-3-8B \citep{grattafiori2024llama} & \url{https://huggingface.co/meta-llama/Meta-Llama-3-8B-Instruct} \\
LLaMA3-70B \citep{grattafiori2024llama} & \url{https://huggingface.co/meta-llama/Llama-3.3-70B-Instruct} \\
GPT-OSS-20B \citep{agarwal2025gpt} & \url{https://huggingface.co/openai/gpt-oss-20b} \\
OpenBioLLM-8B \citep{OpenBioLLMs} & \url{https://huggingface.co/aaditya/Llama3-OpenBioLLM-8B} \\
OpenBioLLM-70B \citep{OpenBioLLMs} & \url{https://huggingface.co/aaditya/Llama3-OpenBioLLM-70B} \\
JSL-MedLlama-3-8B-v2.0 \citep{medllama} & \url{https://huggingface.co/johnsnowlabs/JSL-MedLlama-3-8B-v2.0} \\
HallOumi-8B \citep{oumi2025HallOumi} & \url{https://huggingface.co/oumi-ai/HallOumi-8B} \\
Prometheus-2-8x7B \citep{kim2024prometheus} & \url{https://huggingface.co/prometheus-eval/prometheus-8x7b-v2.0} \\
\hline
\end{tabular}
}
\caption{List of model checkpoints and their download links}
\label{tab:checkpoints}
\end{table*}

\begin{table*}[h]
\centering
\label{tab:models}
\begin{tabular}{lll}
\hline
\textbf{Model} & \textbf{Parameters} \\
\hline
Llama-3-8B    & Temperature = 0.6, Top P = 0.9                                    \\
Llama-3-70B    & Temperature = 0.6, Top P = 0.9                                    \\
Qwen3-8B        & Temperature = 0.7, Top P = 0.8, Top K = 20, Min P = 0      \\
GPT-OSS-20B     & Temperature = 1.0, Top P = 1.0                                    \\
OpenBioLLM-8B & Temperature = 0.0, Top P = 0.9                                   \\
OpenBioLLM-70B & Temperature = 0.0, Top P = 0.9                                   \\
MedLlama3-8B    & Temperature = 0.0, Top P = 0.9 , Top K = -1, Min P = 0           \\
Halloumi-8B     & Temperature = 0.0, Top P = 0.9 , Top K = -1, Min P = 0           \\
Prometheus-2-8x7B & Temperature = 1.0, Top P = 0.9                                  \\
\hline
\end{tabular}
\caption{Models and Configurations}
\label{tab:samplng}
\end{table*}

\section{Sample Size Selection}
\label{app:sample_size}

To ensure a statistically reliable estimate of dataset quality, we determine the sample size using standard methods for proportion estimation (see \citep{cochran1977sampling}). Specifically, we use $n = \frac{Z^2 p (1 - p)}{E^2}$ with a 95\% confidence level ($Z = 1.96$), a conservative estimate $p = 0.5$, and a margin of error $E = 0.05$, yielding $n \approx 385$. This implies that the estimated proportion of correctly labeled samples is within $\pm 5\%$ of the true value with 95\% confidence. Given the large size of the dataset ($\sim 800$k), no finite population correction is applied. To additionally assess inter-annotator agreement between two annotators, we retain this sample size as it satisfies both quality estimation and agreement analysis requirements. In practice, we annotate 400 samples (100 per task, balanced across labels), slightly exceeding the minimum requirement.

\section{Data Leakage}
\label{app:leakage}

To test for potential data leakage and prior exposure, we performed two analyzes to assess whether the evaluated models may have seen portions of MedHal during pretraining. As a control, we also evaluated the three downstream datasets (ACI-NH, Hall-Gen-DI, and MEDEC-UW), which were released after the pretraining of the evaluated models and are therefore unlikely to be contaminated. All analyzes were conducted on random samples of 1,000 examples from each dataset and the MedHal source.

First, following \citet{shi2024detecting}, we assess potential pretraining exposure for each test dataset by computing two metrics on the first 128 tokens of the context: \emph{Min-K\% Prob}, which measures the average surprise of the least likely $K\%$ of tokens, and \emph{PPL}, which measures model perplexity (i.e., average next-token surprise). These two metrics provide an indication of how familiar a model is with the input text. Higher Min-K\% probability and lower perplexity indicate that a model assigns greater likelihood to the input, which may suggest prior exposure.

Tables~\ref{tab:mink} and~\ref{tab:ppl} show no consistent separation between MedHal and the control datasets across models and tasks. In several cases, the control datasets exhibit stronger potential contamination signals than MedHal (e.g., MEDEC-UW has the highest Min-K\% values across all models), while Hall-Gen-DI often has the lowest perplexity. Instead, the observed variation appears more closely related to dataset characteristics, such as task and source text type (e.g., QA versus summarization), than to prior exposure to MedHal.

\begin{table*}[h]
\centering
\caption{Min-K\% Prob. ($K=20$) computed on the first 128 tokens of the context.}
\label{tab:mink}
\resizebox{\textwidth}{!}{
\begin{tabular}{llccccc}
\toprule
Source & Type & HallOumi & OpenBioLLM & Llama-3-8B & Qwen3-8B & Llama-3.3-70B \\
\midrule
QA & MedHal & 8.42 & 8.90 & 7.11 & 8.53 & 7.80 \\
Summarization & MedHal & 7.78 & 8.64 & 7.09 & 7.85 & 7.62 \\
Information Extraction & MedHal & 6.98 & 7.76 & 6.01 & 7.28 & 6.45 \\
NLI & MedHal & 6.23 & 7.06 & 5.42 & 6.26 & 5.85 \\
\midrule
MEDEC-UW & Control & 8.74 & 9.74 & 8.16 & 9.56 & 8.40 \\
ACI-NH & Control & 7.80 & 8.66 & 6.72 & 8.23 & 6.99 \\
Hall-Gen-DI & Control & 6.15 & 7.17 & 5.22 & 6.45 & 5.65 \\
\bottomrule
\end{tabular}
}
\end{table*}

\begin{table*}[h]
\centering
\caption{Median perplexity computed on the first 128 tokens of the context.}
\label{tab:ppl}
\resizebox{\textwidth}{!}{
\begin{tabular}{llccccc}
\toprule
Source & Type & HallOumi & OpenBioLLM & Llama-3-8B & Qwen3-8B & Llama-3.3-70B \\
\midrule
QA & MedHal & 6.8 & 8.7 & 6.2 & 6.8 & 5.9 \\
Summarization & MedHal & 14.5 & 18.6 & 12.0 & 13.8 & 12.0 \\
Information Extraction & MedHal & 10.1 & 12.5 & 8.1 & 10.5 & 8.2 \\
NLI & MedHal & 19.8 & 18.7 & 12.9 & 16.8 & 15.5 \\
\midrule
MEDEC-UW & Control & 26.2 & 39.9 & 21.2 & 32.2 & 20.6 \\
ACI-NH & Control & 13.4 & 17.4 & 10.7 & 13.9 & 10.3 \\
Hall-Gen-DI & Control & 6.1 & 7.6 & 4.6 & 5.9 & 4.3 \\
\bottomrule
\end{tabular}
}
\end{table*}

Second, we use the \emph{Time Travel in LLMs} framework \citet{golchin2023time}, in which we prompt the models to complete the context text either without additional information (\emph{general}) or given the dataset name (\emph{guided}). We then compute the ROUGE-L similarity between the generated completions and the original text. Higher similarity under guided prompting may indicate potential prior exposure, as providing the dataset name could help the model retrieve memorized text.

Table~\ref{tab:timetravel} shows evidence of potential prior exposure only for the QA subset of MedHal, where Qwen3-8B and both Llama models exhibit small but statistically significant increases in ROUGE-L under guided prompting. No comparable pattern is observed for the remaining MedHal sources, and the downstream control datasets likewise show no systematic evidence of prior exposure.

\begin{table*}[h]
\centering
\caption{Time Travel mean ROUGE-L delta (guided $-$ general). Significant cells (bootstrap $p \leq 0.05$) are marked with *.}
\label{tab:timetravel}
\resizebox{\textwidth}{!}{
\begin{tabular}{llccccc}
\toprule
Source & Type & HallOumi & OpenBioLLM & Llama-3-8B & Qwen3-8B & Llama-3.3-70B \\
\midrule
QA & MedHal & $-0.067$ & $-0.001$ & $+0.012^{*}$ & $+0.032^{*}$ & $+0.022^{*}$ \\
Summarization & MedHal & $-0.039$ & $-0.034$ & $-0.084$ & $-0.009$ & $-0.046$ \\
Information Extraction & MedHal & $-0.057$ & $-0.022$ & $-0.001$ & $-0.004$ & $-0.039$ \\
NLI & MedHal & $-0.071$ & $-0.007$ & $-0.004$ & $+0.005$ & $-0.001$ \\
\midrule
MEDEC-UW & Control & $-0.009$ & $+0.005$ & $-0.003$ & $+0.004$ & $-0.048$ \\
ACI-NH & Control & $-0.052$ & $-0.011$ & $+0.000$ & $+0.005$ & $-0.024$ \\
Hall-Gen-DI & Control & $-0.079$ & $-0.024$ & $-0.026$ & $+0.007$ & $-0.013$ \\
\bottomrule
\end{tabular}
}
\end{table*}

Overall, these experiments indicate that prior exposure to some of the datasets making up MedHal is possible for some of the tested models. However, the contamination analyzes do not provide evidence of systematic exposure across MedHal sources, and the observed signals are limited to the QA subset under the Time Travel analysis. Importantly, the analyzes also indicate that the downstream datasets do not exhibit systematic evidence of contamination. The results on these downstream datasets, therefore, provide evidence that the gains obtained through MedHal training extend beyond the training sources and reflect generalization to datasets that are unlikely to have been present during pretraining.

\section{Task Ablation}
\label{task_ablation}
We perform an ablation on the task training data to better understand the individual contribution of each task to the models' overall performance. In this setting, we fine-tune the Llama-3-8B base model separately on the training samples for each task: Summarization, Question Answering, and Information Extraction. The same prompt format is used across all experiments to ensure consistency. We exclude the Natural Language Inference (NLI) task from this analysis because it contains significantly fewer samples compared to the others, which would make a direct comparison of its impact inequitable. By isolating each task, this ablation allows us to assess its relative importance. A substantial performance gain from a single task would indicate its high value, whereas a decrease or negligible improvement compared to the baseline could suggest the task is less critical for the overall goal. Results are shown in Table \ref{tab:results_MedHal_fine_tuned_tasks}.

\begin{table*}[h]
\centering
\begin{tabular}{|c|cc@{\hspace{0.4em}}cc|cc|}
\toprule
\multirow{2}{*}{\textbf{Task}} & \multicolumn{4}{c|}{\textbf{Hallucination Detection}} & \multicolumn{2}{c|}{\textbf{Explanation}} \\ \cline{2-5} \cline{6-7}
 & \textbf{H-F1} & \textbf{NH-F1} & \textbf{Mac-F1} & \textbf{$\Delta$F1} & \textbf{ECS} & \textbf{N} \\ \midrule
Summarization & 0.757 & 0.829 & 0.794 & +0.09 & 0.72 & 20,142 \\ \hline
Question Answering & 0.759 & 0.754 & 0.756 & +0.06 & 0.71 & 19,974 \\ \hline
Information Extraction & 0.758 & 0.732 & 0.745 & +0.05 & 0.68 & 18,013 \\ \bottomrule
\end{tabular}
\caption{Performance of Llama-3-8B on MedHal's test set after fine-tuning using only samples associated to certain tasks.
N refers to the number of samples where an explanation was extracted when prompting the model,
H is for `Hallucinated', NH for `Non-Hallucinated' and ECS for `Embedding Cosine Similarity', $\Delta$F1 is the improvement in Macro-F1 between the base and fine-tuned models.}
\label{tab:results_MedHal_fine_tuned_tasks}
\end{table*}

Table \ref{tab:results_MedHal_fine_tuned_tasks} shows that each task in the dataset contributes uniquely to improving model performance in detecting factual and non-factual content, with all tasks boosting the baseline F1 score by at least 5 Macro-F1 points. Among them, the \textit{Summarization} task shows the most significant impact. Fine-tuning solely on \textit{Summarization} leads to substantially greater gains over the baseline than fine-tuning on \textit{Question Answering} or \textit{Information Extraction}. These improvements are consistent across both "Factuality" and "Explanation" metrics. The effectiveness of \textit{Summarization} can be attributed to its inherent complexity. Unlike \textit{Question Answering}, which typically draws on general medical knowledge, \textit{Summarization} requires a more complex understanding of the context and the statement. While \textit{Information Extraction} also involves contextual analysis, it deals with short, focused statements. \textit{Summarization}, in contrast, involves longer statements and minor inconsistencies, requiring the model to analyze both the statement and the broader context. This might lead to the model learning more robust, generalizable representations for hallucination detection, which also benefit tasks like \textit{Information Extraction} and some \textit{Question Answering} cases. Representations learned from \textit{Question Answering} and \textit{Information Extraction} do not transfer as effectively to other tasks in the dataset.

\section{Training configuration}
\label{app:training_config}
Figure \ref{fig:training_config} presents the QLoRA configuration for all MedHal fine-tuning. All training was done using the computational setup described in Section \ref{app:compute_setup}. Training was done using the TRL \citep{vonwerra2022trl} framework.

\begin{figure*}[h]
\centering
\fbox{%
  \begin{minipage}{0.8\textwidth}
    load\_in\_4bit: true

    max\_seq\_len: 8192

    per\_device\_train\_batch\_size: 8

    per\_device\_eval\_batch\_size: 8

    num\_train\_epochs: 1

    gradient\_accumulation\_steps: 2

    optim: paged\_adamw\_8bit

    r: 16

    lora\_alpha: 16
  \end{minipage}
}
\caption{Training configuration with QLoRA}
\label{fig:training_config}
\end{figure*}

\section{Computational Setup}
\label{app:compute_setup}

All experiments (training and inference) were conducted on a server equipped with 4 NVIDIA H100 80GB GPUs. Training runs took around 100 GPU-hours on average, while inference was highly dependent on the dataset, ranging from 1 GPU-hour for the small downstream datasets, to 12 GPU-hours for the full MedHal test set.

\section{Annotation Guidelines}
\label{app:ann_guidelines}

\noindent{\textbf{Evaluation Criteria}}\\

\underline{What Constitutes a ``Factual'' Statement?}\\

A statement is considered factual if and only if:
\begin{enumerate}
    \item Every piece of information mentioned in the statement can be directly supported by the provided context,  OR
    \item When no context is provided, the statement represents well-established, accurate medical knowledge.
    \item There are no contradictions between the statement and the context.
    \item The statement does not include unsupported claims or extrapolations beyond what the context states.
\end{enumerate}

\textbf{Evaluation Process}
\begin{enumerate}
    \item Review the Context\\
    Carefully read the provided context (e.g., research abstract, clinical information, etc.)

    \item Analyze the Statement\\
    Break down the statement into individual claims and check each claim against the context.

    \item Cross-Reference with Medical Knowledge\\
    For statements without context, evaluate them against established medical facts.

    \item Make Your Assessment\\
    Fill in the evaluation columns:
    \begin{itemize}
        \item \textit{Valid Column:}
        \begin{itemize}
            \item YES: You agree with the current label (TRUE for factual / FALSE for non-factual)
            \item NO: You disagree with the current label
        \end{itemize}
        \item \textit{Comment Column:} Provide your reasoning, including:
        \begin{itemize}
            \item Specific discrepancies found (if any)
            \item Medical knowledge that supports or contradicts the statement
            \item Concerns about accuracy or completeness
        \end{itemize}
    \end{itemize}
\end{enumerate}

\underline{Example Evaluation} \\

\textbf{Context:} \\

\emph{``A randomized controlled trial of 200 patients found that Drug A reduced hospital readmissions by 30\% compared to placebo ($p<0.05$).''}

\vspace{0.5em}
Statement: \\
\emph{``Drug A significantly reduces hospital readmissions by 30\% in all patients.''}

\vspace{0.5em}
Current Label: TRUE

\vspace{0.5em}
Your Evaluation:
\begin{itemize}
    \item \textit{Valid:} NO
    \item \textit{Comment:} ``While the 30\% reduction is accurate, the statement overgeneralizes by claiming effectiveness `in all patients' when the study only included 200 patients with specific characteristics. The word `significantly' is supported by $p<0.05$.''
\end{itemize}

\end{document}